\documentclass[12pt]{article}
\usepackage{setspace} 
\usepackage{graphicx}
\usepackage{xcolor}
\usepackage{authblk}

\usepackage[bottom=2cm, right=2cm, left=1.0cm, top=2cm]{geometry}
\usepackage{hyperref}
\usepackage{cite} 
\usepackage{threeparttable}
\usepackage{algorithm2e}
\RestyleAlgo{ruled}
\usepackage{booktabs}       
\usepackage{multirow}

\title{Low-resource classification of mobility functioning information in clinical sentences using large language models}

\author[1, 2]{Tuan-Dung Le, MS}
\author[1, 2]{Thanh Duong, BS}
\author[2, 3, *]{Thanh Thieu, PhD}

\affil[1]{Department of Computer Science and Engineering,
	University of South Florida, Tampa, FL}
\affil[2]{Department of Machine Learning, Moffitt Cancer Center, Tampa, FL}
\affil[3]{Department of Oncological Sciences, University of South Florida Morsani College of Medicine, Tampa, FL}

\begin{document}

\date{}
\maketitle
\noindent *Corresponding author: Thanh Thieu, PhD, Machine Learning Department, Moffitt Cancer Center and Research Institute, 12902 USF Magnolia Drive, Tampa, FL 33612, USA \\
Email: thanh.thieu@moffitt.org\\


\noindent Word count (excluding title page, abstract, references, figures, and tables): 3540

\pagebreak

\begin{abstract}
\textbf{Objective} \quad Function is increasingly recognized as an important indicator of whole-person health.
This study evaluates the ability of publicly available large language models (LLMs) to accurately identify the presence of functioning information from clinical notes.
We explore various strategies to improve the performance on this task.

\textbf{Materials and Methods} \quad 
We collect a balanced binary classification dataset of 1000 sentences from the Mobility NER dataset, which was curated from n2c2 clinical notes.
For evaluation, we construct zero-shot and few-shot prompts to query the LLMs whether a given sentence contains mobility functioning information.
Two sampling techniques, random sampling and k-nearest neighbor (kNN)-based sampling, are used to select the few-shot examples.
Furthermore, we apply a parameter-efficient prompt-based fine-tuning method to the LLMs and evaluate their performance under various training settings.

\textbf{Results} \quad
Flan-T5-xxl outperforms all other models in both zero-shot and few-shot settings, achieving a F1 score of 0.865 with a single demonstrative example selected by kNN sampling.
In prompt-based fine-tuning experiments, this foundation model also demonstrates superior performance across all low-resource settings, particularly achieving an impressive F1 score of 0.922 using the full training dataset.
The smaller model, Flan-T5-xl, requires fine-tuning with only 2.3M additional parameters to achieve comparable performance to the fully fine-tuned Gatortron-base model, both surpassing 0.9 F1 score.

\textbf{Conclusion} \quad 
Open-source instruction-tuned LLMs demonstrate impressive in-context learning capability in the mobility functioning classification task. 
The performance of these models can be further improved by continuing fine-tuning on a task-specific dataset. 
\end{abstract}
\pagebreak

\section*{Introduction}



Human function refers to the various abilities, capacities, and processes that enable an individual to perform tasks, engage with their environment to meet basic needs and fulfill expected roles in daily life\cite{high2019use}.
In the last few years, functional status has been increasingly recognized as an important health indicator in addition to mortality and morbidity \cite{stucki2017functioning, hopfe2018optimizing}.
The term "function" is defined by the International Classification of Functioning, Disability, and Health (ICF) \cite{icf} as the interaction between an individual's health condition and contextual factors, including both body functions, structures, activities, and participation.
By assessing and documenting human functions within the ICF, healthcare professionals can better comprehend the specific problems that an individual experiences due to health conditions or disabilities. 
Therefore, identifying and understanding functional status information is important in optimizing interventions, treatments, and support regimens that improve overall functioning and quality of life.



Although the volume of healthcare data has increased tremendously in recent years, most functioning information still remains hidden in free text.
This restricts the extraction of valuable real-world insights that could enhance patient care and further research advancements.
However, effective extraction and interpretation of functional status data from electronic health records (EHRs) remains a critical research challenge.
To tackle this, we develop classification methods that can rapidly identify the presence of relevant functional information within clinical notes.
This innovation enables healthcare professionals, including doctors, nurses, and medical annotators to use the model as an assistant, thus dramatically reducing the time to read the entire clinical notes by pinpointing critical sentences for their review.

\subsection*{Objective}
Our contributions are three folds: 

1. We evaluate the performance of publicly available instruction-tuned LLMs in both zero-shot and few-shot settings for the task of identifying the presence of mobility functioning in clinical sentences. 
Our experiments demonstrate that employing k-nearest neighbors (kNN)-based sampling over random sampling can further enhance few-shot performance.

2. We further fine-tune these LLMs employing the parameter-efficient prompt-based fine-tuning approach, which yields higher classification accuracy than the few-shot approach. 
Analyzing this approach under various low-resource settings, we demonstrates that it also outperforms the traditional full-model fine-tuning approach on BERT-based models.

3. We release the annotated dataset used in this study to the research community, enabling further applications of LLMs in healthcare.

\subsection*{Background}

\subsubsection*{Functional Status Information (FSI)}

Automatically extracting and coding functional status information (FSI) from clinical text is still a nascent field in clinical natural language processing (NLP).
Early attempts to identify FSI relied heavily on manual annotation by clinical staff \cite{kuang2015representation, mahmoud2014icf, greenwald2017novel, skube2018characterizing} or automated method restricted to specific ICF codes \cite{kukafka2006human}.
To advance this research area, Thieu et al \cite{thieu2017inductive, thieu2021comprehensive} constructed a private dataset focusing on extracting Mobility-related FSI from physical therapy notes at the National Institutes of Health (NIH).
They focused on Mobility domain of the ICF due to its well-defined and observable nature as a construct of human functioning.
They developed named-entity recognition (NER) models to extract Mobility-related entities and achieved impressive average F1 score of 84.90\%.
However, to obtain such high performance, extensive domain expert annotation of clinical notes is required.
Furthermore, the privacy of the dataset restricts access and follow-up studies \cite{newman-griffis-zirikly-2018-embedding, newman2019classifying, newman2021linking} to a handful of NIH researchers.
To mitigate these limitations, Le et al. \cite{le2023leveraging} proposed an active learning pipeline that strategically annotates examples from n2c2 research datasets, aiming to create a publicly available Mobility NER dataset with less annotation effort.
However, they found that Mobility-related entities in public resources such as the n2c2 clinical notes were scarce and data imbalance resulted in lower NER performance.
In this paper, we identify the presence of mobility functioning information in clinical notes at sentence-level rather than entity-level.
This allows us to leverage the power of LLMs and achieve high classification accuracy with limited amount of data.

\subsubsection*{Large language models (LLMs)}

Over the last few years, LLMs \cite{brown2020language, rae2021scaling, tay2022ul2, chowdhery2023palm} pre-trained on massive text corpora have shown strong performance in a variety of NLP tasks. 
Furthermore, instruction tuning \cite{sanh2021multitask, chung2022scaling} and reinforcement learning from human feedback (RLHF) \cite{ouyang2022training} are helping LLMs better understand instructions and align with human preferences.
Instruction-tuned models like Med-PaLM 2 \cite{singhal2023towards} and GPT-4 \cite{OpenAI2023, nori2023capabilities} show remarkable performance on medical competency examinations and benchmark datasets.
However, using these proprietary models for other clinical NLP tasks requires sending medical data through web API, raising privacy concerns of leaking patient data.
Smaller and open-source instruction-tuned models such as T0 \cite{sanh2021multitask}, Flan-T5 \cite{chung2022scaling} and Llama 2 \cite{touvron2023llama} are better fit for this purpose. 
Their smaller size, faster in-context learning inference, and the ability to further improve with parameter-efficient fine-tuning (PEFT)\cite{hu2021lora, li2021prefix, lester2021power} make them ideal for broader clinical NLP applications.

\section*{MATERIALS AND METHODS}

\subsection*{Dataset}


We collect a small, high-quality gold standard dataset by drawing annotated, adjudicated sentences from the Mobility named-entity recognition (NER) dataset \cite{le2023leveraging}.
Even though the Mobility NER data set is large, only a small number of sentences are both double-blind annotated and further adjudicated by a third, senior annotator.
The dataset contains sentences derived from clinical notes in the n2c2 NLP research corpus \cite{n2c2portal}, in which each sentence is annotated with four types of entities: Action, Mobility, Assistance, and Quantification.

To generate labels for our mobility classification dataset, we mark sentences containing at least one Mobility entity as positive, while those without any Mobility entity as negative.
We select a five-week subset of the NER dataset and collect positive sentences within this subset. 
This five-week subset has higher annotation quality than the rest of the NER dataset by incorporating the adjudication of a senior expert annotator.
We exclude sentences that exceed 45 tokens in length because these sentences are often incomplete text fragments poorly segmented from the original clinical notes.
The final gold standard dataset contains 500 positive sentences.

Because a classification task requires negative sentences, we retain all negative sentences within the five-week subset.
However, the number of negative sentences is still significantly less than the number of positive sentences. 
To balance the data set, we randomly collect additional negative sentences from the remaining NER dataset outside the five-week subset until it reaches 500 negative sentences.

\begin{figure}[t]
    \centering
    \includegraphics[width=1\textwidth]{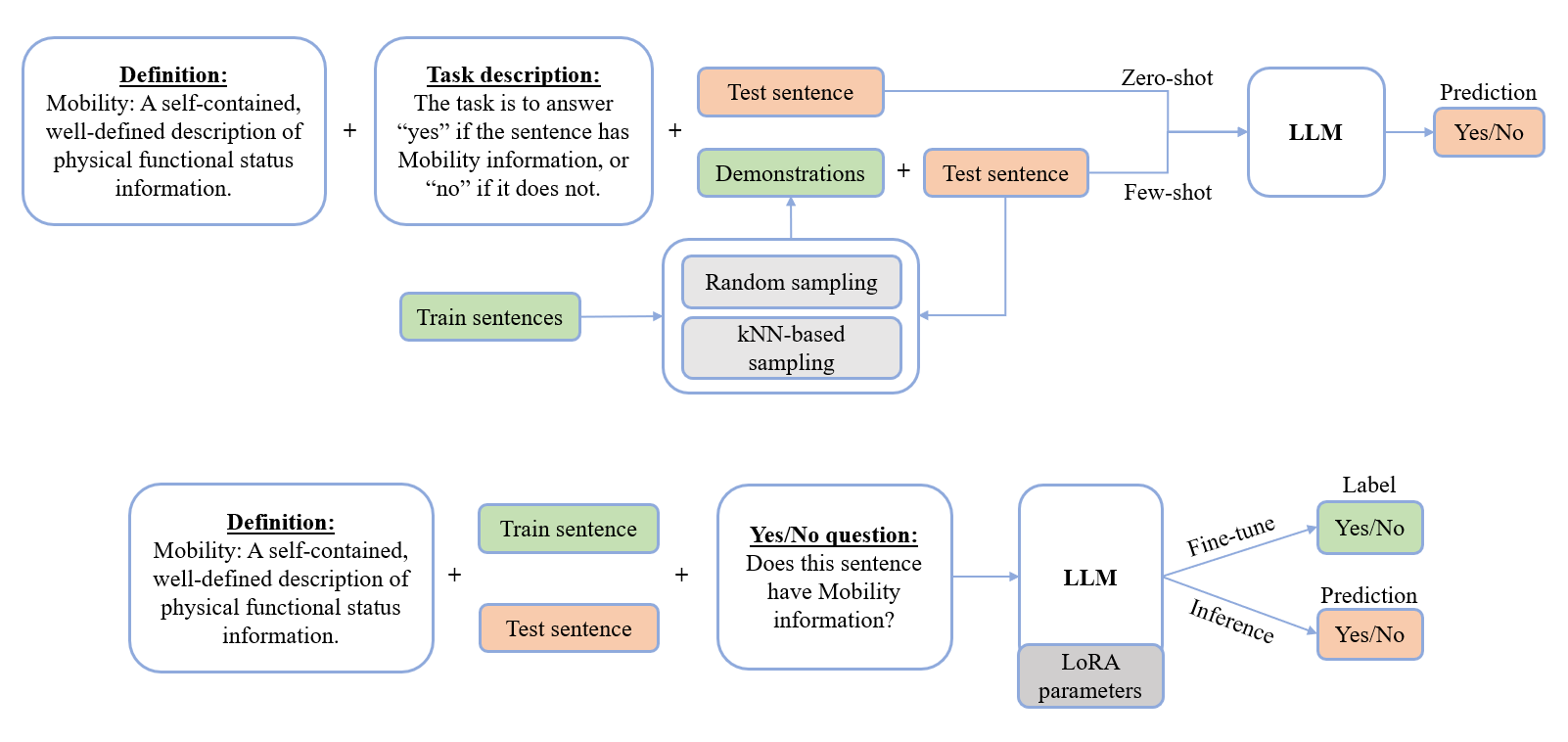}
    \caption{Top: In-context learning. Bottom: Parameter-efficient prompt-based fine-tuning.}
    \label{fig:fig1}
\end{figure}

\subsection*{In-context Learning}


We follow the standard prompt-based in-context learning approach, as described in Brown et al.\cite{brown2020language}.
Given an input sentence $x$, we construct a prompt and feed it to the LLM to generate an answer $y$.
In our experiment, we ask the LLM to determine whether a given sentence contains functioning information ($y \in \{yes, no \}$) 

\subsubsection*{Prompt Construction}

Our prompt contains three main components: task instruction, demonstrations, and test sentence.

Task instruction first provides the LLM with context regarding mobility information, then guides it to answer the question directly and in the expected format.
Specifically, we start the task instruction with the definition of Mobility entity as defined in the NER formulation \cite{thieu2017inductive, thieu2021comprehensive}: \textit{Mobility: A self-contained, well-defined description of physical functional status information}. 
The following sentence describes the classification task: \textit{The task is to answer yes if the sentence contains Mobility information, or no if it does not}.

Demonstrations consist of $k$ annotated examples, each containing a sentence paired with its corresponding label: (Sentence: $x^1_d$, Answer: $y^1_d$), ..., (Sentence: $x^k_d$, Answer: $y^k_d$)
These demonstrations serve as examples for the LLM to contemplate in decision-making and illustrate how to follow the task instructions to generate answers in the expected format.
It is important to note that demonstrations are required for the few-shot learning setup, but not for the zero-shot learning setup.

Test sentence is appended at the end of the prompt in the following format: \textit{Sentence: $x_{test}$ Answer: [LLM will generate the answer here]}

\subsubsection*{Sampling Few-shot Demonstrations}

The few-shot setup requires demonstrations drawn from the training dataset.
As observed by Zhao et al.\cite{pmlr-v139-zhao21c}, class imbalance within a text classification prompt can lead to biased LLM predictions due to the influence of the majority class.
To mitigate this issue, we sample an equal number of positive and negative sentences for demonstrations in even shot scenarios $(k= 2, 4, 8)$.
However, for one-shot scenarios (k=1), we only use a positive sentence. 
Our preliminary experiments indicated that using a single positive example yielded a higher accuracy than using a single negative example or a randomly selected example.
Additionally, we apply two straightforward strategies for selecting demonstration examples: Random sampling and kNN-based sampling.
\begin{itemize}
    \item \textbf{Random sampling} \quad 
    We randomly, uniformly sample $k$ examples from the training set.
    \item \textbf{kNN-based sampling} \quad 
    Liu et al.\cite{liu-etal-2022-makes} observed that the performance of LLMs is sensitive to the choice of in-context examples, indicating that random sampling strategy may not consistently produce reliable results.
    Instead, they suggest retrieving examples that are semantically similar to a test sentence to create more effective prompts for the LLM.
    Following their approach, we utilize sentence embeddings to select examples that are close to the input in the embedding space.
    Specifically, we use the all-mpnet-base-v2 model developed by Reimers et al.\cite{reimers-2019-sentence-bert}, which was pre-trained on an extensive dataset of over 1 billion sentence pairs, to generate embeddings for all training sentences.
    For each test sentence, we first generate its corresponding embedding and calculate its cosine similarity with each pre-computed training sentence embedding to find the kNN examples. 
\end{itemize}

\subsection*{Low-resource Fine-tuning}

Instruction tuning has demonstrated its effectiveness in improving the overall performance of LLMs across a wide range of unseen tasks \cite{sanh2021multitask, chung2022scaling}.
Furthermore, the performance of instruction-finetuned LLMs on a single, specific downstream task can be further enhanced by continuous fine-tuning on task-specific data.
This raises the question of how much task-specific data is required for fine-tuning to surpass the zero-shot and few-shot performance of instruction-tuned LLMs.
To answer this question, we implement a parameter-efficient prompt-based fine-tuning method and train it using different numbers of training examples. 
This approach requires one or more GPUs depending on the LLM's size.
For comparison, we also employ traditional full-model fine-tuning on a smaller language model, specifically a masked language model (MLM), which can be effectively trained on a single GPU with limited memory.

\subsubsection*{Parameter-efficient fine-tuning}

\emph{Prompt-based fine-tuning}:
We formulate the classification task as a next token prediction task.
Given a prompt, the model predicts whether the next token should be 'yes' or 'no', indicating the presence or absence of mobility information in the input sentence. 
Specifically, we construct the prompt by concatenating the definition of mobility with the input sentence and a direct yes/no question: "Does this sentence have Mobility information?"

\emph{Low Rank Adaptation}:
Fully fine-tuning a LLM is computationally demanding, requiring significant training time and GPU resources.
PEFT techniques \cite{hu2021lora, li2021prefix, lester2021power, peft} have been widely used to alleviate this problem. 
By selectively optimizing only a small number of parameters, these methods significantly reduce computational and storage requirements. 
We employ a PEFT method called Low Rank Adaptation (LoRA) \cite{hu2021lora}. 
This method introduces rank-decomposition weight matrices to the existing model weights and trains only these newly added weights.
In our experiments, we apply LoRA with rank $r=2$ on all weight matrices in the self-attention module, including $W_{query}$, $W_{key}$, $W_{value}$ and $W_{output}$.
Next, we fine-tune the model for 10 epochs with an initial learning rate of 1e-3.

\subsubsection*{Full model fine-tuning}

A traditional approach to fine-tuning a large language model for classification tasks involves adding a classification head to a pre-trained, foundational masked language model.
The classification head takes the vector representation of the special [CLS] token as input and generates a logit vector, where each logit represents the likelihood of the input belonging to a particular class. 
A softmax function is then applied to obtain the final class probability prediction.
We fine-tune the model for 10 epochs with an initial learning rate of 5e-5, a lower learning rate compared to the PEFT approach. 
Given that this fine-tuning approach modifies all model parameters, employing a lower learning rate helps prevent catastrophic forgetting and performance degradation.

\subsection*{Evaluation}

LLMs may stochastically generate answers that do not fall into the "yes" or "no" category.
In these cases, we convert such unexpected answers to "no" for consistency. 
While this conversion affects the accuracy metric, it does not affect the F1 score.
Therefore, we report the F1 score as a reliable metric of LLM performance.
For all experiments, we adopt the four-fold cross-validation and calculate a final F1 score as average of F1 scores across the folds.

To investigate the impact of the size of the training dataset on downstream models, we use three training set sizes: 50, 100, and 250 examples. 
Specifically, we randomly select 50 and 100 examples from the training dataset while ensuring that the classes are balanced.
Using 250 examples is equivalent to using the entire training dataset.

\subsection*{Pre-trained Model Selection}

For in-context learning, we utilize publicly available, instruction-tuned foundation LLMs to classify sentences for Mobility information relevancy:
\begin{itemize}
    \item \textbf{T0\cite{sanh2021multitask}:} \quad A T5-based encoder-decoder model fine-tuned on multi-task mixture of natural language instruction datasets to improve the model's zero-shot generalization capabilities.
    We employ T0++ (11B), the most powerful variant of T0, primarily because it is trained on the most extensive collection of instruction datasets.
    \item \textbf{Flan-T5\cite{chung2022scaling}:} \quad Similar to T0, Flan-T5 is derived from T5 but it is fine-tuned on a collection of 1,836 tasks with a variety of instruction templates, including zero-shot, few-shot and chain-of-thought prompts.
    The Flan-T5 series offers a variety of models with parameter sizes ranging from 80M to 11B. 
    For our experiments, we use the Flan-T5-xl (3B) and Flan-T5-xxl (11B) variants.
  
    \item \textbf{Flan-UL2\cite{tay2022ul2}:} \quad A 20B parameter checkpoint of the UL2 model fine-tuned on the Flan dataset with mixture-of-denoisers objective including diverse span corruption and prefix language modeling tasks.
    \item \textbf{Llama 2-Chat\cite{touvron2023llama}:} A dialogue-optimized version of the autoregressive Llama 2 model.
    To improve its ability to follow user instructions, the base Llama 2 model is first fine-tuned using the Flan dataset and an additional annotated dataset.
    The model is then further refined using RLHF to better align with human preferences and enhance its safety features. 
    For our experiments, we use all three available Llama 2-Chat models, ranging in size from 7B to 70B.
\end{itemize}


For low-resource fine-tuning, we also employ smaller, domain-specific language models pre-trained on biomedical and clinical texts:
\begin{itemize}
    \item \textbf{DischargeSummaryBERT\cite{alsentzer2019publicly}:} \quad A masked language model (MLM) fine-tuned from 110M-parameter BioBERT using only MIMIC-III discharge-summaries.
    \item \textbf{Gatortron\cite{yang2022large}:} \quad A MLM pre-trained on a combined corpus comprising the University of Florida Health clinical dataset and a subset of MIMIC-III, PubMed, and Wikipedia.
    While there are two publicly available Gatortron models, we only employ the Gatortron-base model with 355M parameters due to the inefficiency of fine-tuning the full Gatortron-medium model, which scales up to 3.9B parameters.
\end{itemize}

\section*{RESULTS}

\subsection*{In-context learning}

\begin{table}[t]
    \centering\caption{Zero-shot and few-shot prompting results. Note: R = Random sampling, K = kNN-based sampling}
    \centering
    \begin{tabular}{p{2.6cm}p{1.8cm}p{1.0cm}p{1.0cm}p{1.0cm}p{1.0cm}p{1.0cm}p{1.0cm}p{1.0cm}p{1.0cm}p{1.0cm}}
        \toprule
        \multirow{2}{*}{Model}
        & \multirow{2}{*}{\#Params}
        & \centering zero 
        & \multicolumn{2}{c}{1 shot} 
        & \multicolumn{2}{c}{2 shot} 
        & \multicolumn{2}{c}{4 shot} 
        & \multicolumn{2}{c}{8 shot}\\
        \cmidrule(lr){4-5} \cmidrule(lr){6-7}\cmidrule(lr){8-9}\cmidrule(lr){10-11}
        
        & \centering 
        & \centering shot
        & \centering R & \centering K
        & \centering R & \centering K
        & \centering R & \centering K
        & \centering R & \centering K \cr
        \midrule 
        T0++ & \centering 11B & 0.698 & 0.514 & \textbf{0.729} & 0.582 & 0.671 & 0.633 & 0.65 & 0.622 & 0.596 \\
        Flan-T5-xl & \centering 3B & 0.775 & 0.765 & \textbf{0.816} & 0.767 & \centering 0.8 & \centering 0.78 & 0.803 & 0.773 & 0.803 \\
        Flan-T5-xxl & \centering 11B & \textbf{0.832} & 0.832 & \textbf{0.865} & 0.828 & 0.858 & 0.839 & 0.851 & 0.831 & 0.848 \\
        Flan-UL2 & \centering 20B & 0.804 & 0.791 & \textbf{0.852} & 0.804 & 0.838 & 0.814 & 0.833 & 0.805 & \centering 0.83 
        \cr
        \midrule 
         & \centering 7B & 0.655 & 0.088 & 0.121 & 0.602 & 0.637 & \textbf{0.663} & 0.657 & 0.476 & 0.549 \\
        Llama 2-Chat & \centering 13B & 0.735 & 0.595 & 0.693 & \centering \textbf{0.74} & 0.723 & 0.728 & 0.722 & 0.694 & 0.711 \\
         & \centering 70B & 0.743 & 0.703 & 0.763 & \textbf{0.765} & 0.764 & 0.755 & 0.764 & 0.76 & 0.764 \\   
        \midrule 
        \midrule 
        \multicolumn{3}{c}{Average improvement K - R}  & \multicolumn{2}{c}{\textbf{+0.079}} & \multicolumn{2}{c}{+0.029} & \multicolumn{2}{c}{+0.01} & \multicolumn{2}{c}{+0.02} \\
        \bottomrule
    \end{tabular}
    \label{table:1} 
\end{table}
Table \ref{table:1} presents performance of the LLMs evaluated on zero-shot and few-shot experiments. 
The zero-shot results show that Flan-t5-xxl outperforms all other models, including the larger Flan-UL2, 13B and 70B Llama 2 Chat models, achieving the highest F1 score of 0.832.
Surprisingly, despite being specifically fine-tuned for zero-shot tasks, T0++ exhibits lower zero-shot performance on our task compared to both the Llama 2 and Flan models.

We observe that increasing the number of demonstration examples does not guarantee improvement in classification.
In few-shot experiments with random sampling, 17 out of 28 experiments show a decrease in performance compared to the zero-shot setting, while the remaining experiments show marginal improvements.
These performance drops are expected for the T0++ model, as it is not trained on few-shot instruction data.
Additionally, Llama 2 Chat 7B model tends to predict 'no' for most test sentences in the 1-shot setting, resulting in a significantly low F1 score.

In 23 out of 28 few-shot settings, kNN-based sampling improves classification performance over random sampling.
However, the performance gains from kNN-based sampling reduce as the number of demonstration examples in the prompt increases. 
Average performance gain reaches 0.08 in 1-shot setting, but decreases to 0.02 in the 8-shot setting.
Interestingly, using one single positive example that is semantically similar to the test sentence gives the best performance for T0 and Flan models. 
Among these models, Flan-T5-xxl achieves the highest F1 score of 0.865.
In contrast, Llama 2 Chat 13B and 70B models obtain their best performance using random sampling in 2-shot setting.

\subsection*{Low-resource Fine-tuning}

Table \ref{table:2} presents the results of fine-tuning experiments conducted under three different low-resource training settings.
Flan-T5-xxl consistently demonstrates superior performance across all three settings, achieving an outstanding F1 score of 0.922 using prompt-based fine-tuning with 250 training examples.
In addition, fine-tuning T0 and Flan-T5-xl with 50 training examples yields significant improvements compared to their few-shot results. 
However, Flan-T5-xxl and Llama 2 Chat 70B exhibit only marginal gains under similar fine-tuning conditions.

Last but not least, Flan-T5-xl prompt-based fine-tuned model achieves comparable performance to the fully fine-tuned Gatortron-base model in the 250 training examples setting, reaching an impressive F1 score of over 0.9.
It should be noted that Flan-T5-xl only requires 2.3M additional LoRA parameters for fine-tuning.


\begin{table}[t]
    \centering\caption{Low-resource fine-tuning results}
    \centering
    \begin{tabular}{ccccccc}
        \toprule
        \multirow{2}{*}{Approach}
        & \multirow{2}{*}{Model}
        & \multirow{2}{*}{\#Params}
        & \multirow{2}{*}{\parbox{2.5cm}{\centering\#Fine-tuned params}}
        & \multicolumn{3}{c}{\#Training examples}
        \cr
        \cmidrule(lr){5-7}
        & & & & 50 & 100 & 250 \cr
        \midrule 
         & T0++ & 11B & 4.7M & 0.8 & 0.875 & 0.916\\
        & Flan-T5-xl & 3B & 2.3M & 0.852 & 0.888 & 0.903\\
        \multirow{3}{*}{\parbox{3.45cm}{\centering Parameter-efficient prompt-based  fine-tuning}} & Flan-T5-xxl & 11B & 4.7M &\textbf{0.871} & \textbf{0.893} & \textbf{0.922}\\    
         & Flan-UL2 & 20B & 6.7M & 0.838 & 0.886 & 0.9\\    
        \cmidrule(lr){2-7}
         &  & 7B & 1.5M & 0.623 & 0.772 & 0.793\\  
         & Llama 2-Chat & 13B & 2.5M & 0.714 & 0.808 & 0.86\\    
         &  & 70B & 5.5M & 0.774 & 0.792 & 0.871\\  
        \midrule
        \centering Full & DischargeSummaryBERT & 110M & 110M & 0.757 & 0.809 & 0.889\\    
        \centering head fine-tuning & Gatortron-base  & 355M & 355M & 0.793 & 0.822 & 0.905\\    
        \bottomrule
    \end{tabular}
    \label{table:2}
\end{table}

\section*{DISCUSSION}

Leveraging the power of LLMs, we have achieved high accuracy in identifying mobility functioning information in clinical notes.

The results show that Flan-T5-xxl consistently outperforms all other models in both in-context learning and prompt-based fine-tuning experiments.
Compared to the similar-size model T0pp, Flan-T5-xxl is pre-trained on a larger dataset that includes T0 training data, which explains its superior performance.
In addition, Flan-UL2 and Llama 2 Chat 70B models achieve lower F1 scores despite having more parameters than Flan-T5-xxl.
Since Flan-UL2 and Flan-T5-xxl are based on the T5 architecture and fine-tuned using the same Flan dataset, we hypothesize that the UL2 mixture of denoisers objective may hurt the performance on our downstream task.
Llama 2 Chat was initially fine-tuned on the Flan dataset before being fine-tuned on its own annotated dataset specifically designed to enhance its ability to handle dialogue-style instructions.
Furthermore, reinforcement learning from human feedbacks (RLHF) was applied to align model behavior with human preferences and instruction following.
Llama 2 Chat's training procedure is optimized for dialogue task, which potentially harms its robustness in our classification task.

A common assumption is that in-context learning performs better when the number of demonstration examples increases.
However, our few-shot experiments using random sampling reveal only marginal improvement over the zero-shot setup, when the number of demonstration examples increases from 1-shot to 8-shot.
These results align with recent observations by Zhao et al. \cite{zhao2023dynamic}, suggesting that the optimal number of demonstration examples may differ depending on the specific task, dataset, and model used.
Furthermore, the use of kNN-based sampling yields better performance than random sampling.
This approach allows the model to take advantage of the task-specific evidence from the entire labeled dataset by employing the most relevant demonstrations.

The parameter-efficient prompt-based fine-tuning allows us to fine-tune LLMs to further improve their performance on our mobility functioning classification task. 
This method requires significantly less training data than the standard full-model fine-tuning on BERT-based models while maintaining better classification accuracy.
This makes it particularly valuable for NLP tasks in healthcare, where manually annotated datasets are scarce and expensive to create.
In low-resource settings with only 50 and 100 examples, Flan-T5-xl 3B achieves second-best performance, only trailing its larger version, Flan-T5-xxl.
The remarkable accuracy is achievable by fine-tuning 2.3M additional parameters, making it possible to train on a single GPU with 16GB memory.
By storing these extra parameters separately as an adapter, we can plug in multiple adapters to a single pre-trained LLM for different classification tasks.

Our study has certain limitations. 
The effectiveness of in-context learning is highly dependent on the quality of the prompts used to steer the LLMs, thus introducing the need for prompt engineering. 
Crafting a high-performance prompt often requires careful consideration of the prompt format \cite{reynolds2021prompt, min2022rethinking}, the selection of demonstration examples \cite{liu-etal-2022-makes}, and the order of these demonstrations \cite{lu2021fantastically, pmlr-v139-zhao21c}.
In our study, we manually create a straightforward prompt comprised of the definition of Mobility as its context, plus a clear task description, and then concatenated with demonstrations and a test sentence in standard format.
However, automatic prompt optimization methods such as APE \cite{zhou2022large} could further improve task performance.
To select a good set of demonstrations, we apply kNN-based sampling and also balance the number of positive and negative demonstrations in a prompt to mitigate the effect of the majority label bias problem \cite{pmlr-v139-zhao21c}.
However, the ordering of the demonstrations in the prompt is not thoroughly investigated in our experiments due to compute limitation.
This might lead to recency bias \cite{pmlr-v139-zhao21c}, where the LLM's predictions favor the class presented near the end of the prompt.
In addition, generative LLMs relying solely on next-token probability can bias toward "yes" or "no" answers due to the influence of the prompt and their intrinsic biases.
Applying contextual calibration techniques \cite{pmlr-v139-zhao21c} could mitigate these biases and improve prediction accuracy.
Lastly, our prompt-based fine-tuning approach is evaluated using a single, hand-crafted prompt similar to the in-context learning approach.

\section*{CONCLUSION}

In conclusion, publicly available instruction-tuned LLMs exhibit impressive in-context learning capabilities to identify the presence of mobility functioning information in clinical notes.
Parameter-efficient prompt-based fine-tuning methods further improve the performance over in-context learning by using a limited amount of training data.
These findings highlight the potential of LLM to tackle NLP tasks in the clinical domain where large, manually annotated datasets are scarce or non-existent, paving the way for broader healthcare applications.

\section*{AUTHOR CONTRIBUTIONS}
TDL conducted all experiments. TDL and TD wrote the manuscript. TT supervised the project and revised the manuscript. 
All authors approved the submitted version.

\section*{ACKNOWLEDGEMENTS}
We would like to thank Zhuqi Miao (ZM), Brittani Smith (BS), and Samuel Alvarado (SA) for curating the Mobility classification dataset.

\section*{FUNDING} 

\section*{CONFLICT OF INTEREST STATEMENT}
The authors do not have conflicts of interest related to this study.

\section*{DATA AVAILABILITY}
Our low-resource Mobility classification dataset will be released on our research group website: \url{https://lailab.info/research} $\rightarrow$ Whole-person Function $\rightarrow$ Data Source $\rightarrow$ View Data Source, or via n2c2's Community Annotations Downloads section.
The n2c2 research datasets are available at \url{https://portal.dbmi.hms.harvard.edu/projects/n2c2-nlp/} to researchers who sign the NLP Research Purpose and Data Use Agreement form.

\bibliographystyle{unsrt}
\bibliography{references}

\begin{thebibliography}{10}

\bibitem{high2019use}
Kevin~P High, Susan Zieman, Jerry Gurwitz, Carl Hill, Jennifer Lai, Thomas Robinson, Mara Schonberg, and Heather Whitson.
\newblock Use of functional assessment to define therapeutic goals and treatment.
\newblock {\em Journal of the American Geriatrics Society}, 67(9):1782--1790, 2019.

\bibitem{stucki2017functioning}
Gerold Stucki and Jerome Bickenbach.
\newblock Functioning: the third health indicator in the health system and the key indicator for rehabilitation.
\newblock {\em European journal of physical and rehabilitation medicine}, 53(1):134--138, 2017.

\bibitem{hopfe2018optimizing}
Maren Hopfe, Birgit Prodinger, Jerome~E Bickenbach, and Gerold Stucki.
\newblock Optimizing health system response to patient’s needs: an argument for the importance of functioning information.
\newblock {\em Disability and rehabilitation}, 40(19):2325--2330, 2018.

\bibitem{icf}
WHO.
\newblock International classification of functioning, disability, and health : Icf. geneva: World health organization, 2001.

\bibitem{kuang2015representation}
Jinqiu Kuang, April~F Mohanty, VH~Rashmi, Charlene~R Weir, Bruce~E Bray, and Qing Zeng-Treitler.
\newblock Representation of functional status concepts from clinical documents and social media sources by standard terminologies.
\newblock In {\em AMIA Annual Symposium Proceedings}, volume 2015, page 795. American Medical Informatics Association, 2015.

\bibitem{mahmoud2014icf}
Rehab Mahmoud, Nashwa El-Bendary, Hoda~MO Mokhtar, and Aboul~Ella Hassanien.
\newblock Icf based automation system for spinal cord injuries rehabilitation.
\newblock In {\em 2014 9th International Conference on Computer Engineering \& Systems (ICCES)}, pages 192--197. IEEE, 2014.

\bibitem{greenwald2017novel}
Jeffrey~L Greenwald, Patrick~R Cronin, Victoria Carballo, Goodarz Danaei, and Garry Choy.
\newblock A novel model for predicting rehospitalization risk incorporating physical function, cognitive status, and psychosocial support using natural language processing.
\newblock {\em Medical care}, 55(3):261--266, 2017.

\bibitem{skube2018characterizing}
Steven~J Skube, Elizabeth~A Lindemann, Elliot~G Arsoniadis, Mari Akre, Elizabeth~C Wick, and Genevieve~B Melton.
\newblock Characterizing functional health status of surgical patients in clinical notes.
\newblock {\em AMIA Summits on Translational Science Proceedings}, 2018:379, 2018.

\bibitem{kukafka2006human}
Rita Kukafka, Michael~E Bales, Ann Burkhardt, and Carol Friedman.
\newblock Human and automated coding of rehabilitation discharge summaries according to the international classification of functioning, disability, and health.
\newblock {\em Journal of the American Medical Informatics Association}, 13(5):508--515, 2006.

\bibitem{thieu2017inductive}
Thanh Thieu, Jonathan Camacho, Pei-Shu Ho, Julia Porcino, Min Ding, Lisa Nelson, Elizabeth Rasch, Chunxiao Zhou, Leighton Chan, Diane Brandt, et~al.
\newblock Inductive identification of functional status information and establishing a gold standard corpus: A case study on the mobility domain.
\newblock In {\em 2017 IEEE International Conference on Bioinformatics and Biomedicine (BIBM)}, pages 2319--2321. IEEE, 2017.

\bibitem{thieu2021comprehensive}
Thanh Thieu, Jonathan~Camacho Maldonado, Pei-Shu Ho, Min Ding, Alex Marr, Diane Brandt, Denis Newman-Griffis, Ayah Zirikly, Leighton Chan, and Elizabeth Rasch.
\newblock A comprehensive study of mobility functioning information in clinical notes: entity hierarchy, corpus annotation, and sequence labeling.
\newblock {\em International journal of medical informatics}, 147:104351, 2021.

\bibitem{newman-griffis-zirikly-2018-embedding}
Denis Newman-Griffis and Ayah Zirikly.
\newblock Embedding transfer for low-resource medical named entity recognition: A case study on patient mobility.
\newblock In {\em Proceedings of the {B}io{NLP} 2018 workshop}, pages 1--11, Melbourne, Australia, July 2018. Association for Computational Linguistics.

\bibitem{newman2019classifying}
Denis Newman-Griffis, Ayah Zirikly, Guy Divita, and Bart Desmet.
\newblock Classifying the reported ability in clinical mobility descriptions.
\newblock {\em arXiv preprint arXiv:1906.03348}, 2019.

\bibitem{newman2021linking}
Denis Newman-Griffis, Jonathan Camacho~Maldonado, Pei-Shu Ho, Maryanne Sacco, Rafael Jimenez~Silva, Julia Porcino, and Leighton Chan.
\newblock Linking free text documentation of functioning and disability to the icf with natural language processing.
\newblock {\em Frontiers in rehabilitation sciences}, 2:742702, 2021.

\bibitem{le2023leveraging}
Tuan-Dung Le, Zhuqi Miao, Samuel Alvarado, Brittany Smith, William Paiva, and Thanh Thieu.
\newblock Leveraging deep active learning to identify low-resource mobility functioning information in public clinical notes.
\newblock {\em arXiv preprint arXiv:2311.15946}, 2023.

\bibitem{brown2020language}
Tom Brown, Benjamin Mann, Nick Ryder, Melanie Subbiah, Jared~D Kaplan, Prafulla Dhariwal, Arvind Neelakantan, Pranav Shyam, Girish Sastry, Amanda Askell, et~al.
\newblock Language models are few-shot learners.
\newblock {\em Advances in neural information processing systems}, 33:1877--1901, 2020.

\bibitem{rae2021scaling}
Jack~W Rae, Sebastian Borgeaud, Trevor Cai, Katie Millican, Jordan Hoffmann, Francis Song, John Aslanides, Sarah Henderson, Roman Ring, Susannah Young, et~al.
\newblock Scaling language models: Methods, analysis \& insights from training gopher.
\newblock {\em arXiv preprint arXiv:2112.11446}, 2021.

\bibitem{tay2022ul2}
Yi~Tay, Mostafa Dehghani, Vinh~Q Tran, Xavier Garcia, Jason Wei, Xuezhi Wang, Hyung~Won Chung, Dara Bahri, Tal Schuster, Steven Zheng, et~al.
\newblock Ul2: Unifying language learning paradigms.
\newblock In {\em The Eleventh International Conference on Learning Representations}, 2022.

\bibitem{chowdhery2023palm}
Aakanksha Chowdhery, Sharan Narang, Jacob Devlin, Maarten Bosma, Gaurav Mishra, Adam Roberts, Paul Barham, Hyung~Won Chung, Charles Sutton, Sebastian Gehrmann, et~al.
\newblock Palm: Scaling language modeling with pathways.
\newblock {\em Journal of Machine Learning Research}, 24(240):1--113, 2023.

\bibitem{sanh2021multitask}
Victor Sanh, Albert Webson, Colin Raffel, Stephen~H Bach, Lintang Sutawika, Zaid Alyafeai, Antoine Chaffin, Arnaud Stiegler, Teven~Le Scao, Arun Raja, et~al.
\newblock Multitask prompted training enables zero-shot task generalization.
\newblock {\em arXiv preprint arXiv:2110.08207}, 2021.

\bibitem{chung2022scaling}
Hyung~Won Chung, Le~Hou, Shayne Longpre, Barret Zoph, Yi~Tay, William Fedus, Yunxuan Li, Xuezhi Wang, Mostafa Dehghani, Siddhartha Brahma, et~al.
\newblock Scaling instruction-finetuned language models.
\newblock {\em arXiv preprint arXiv:2210.11416}, 2022.

\bibitem{ouyang2022training}
Long Ouyang, Jeffrey Wu, Xu~Jiang, Diogo Almeida, Carroll Wainwright, Pamela Mishkin, Chong Zhang, Sandhini Agarwal, Katarina Slama, Alex Ray, et~al.
\newblock Training language models to follow instructions with human feedback.
\newblock {\em Advances in Neural Information Processing Systems}, 35:27730--27744, 2022.

\bibitem{singhal2023towards}
Karan Singhal, Tao Tu, Juraj Gottweis, Rory Sayres, Ellery Wulczyn, Le~Hou, Kevin Clark, Stephen Pfohl, Heather Cole-Lewis, Darlene Neal, et~al.
\newblock Towards expert-level medical question answering with large language models.
\newblock {\em arXiv preprint arXiv:2305.09617}, 2023.

\bibitem{OpenAI2023}
OpenAI.
\newblock Gpt-4 technical report.
\newblock {\em arXiv preprint arXiv:2303.08774}, 2023.

\bibitem{nori2023capabilities}
Harsha Nori, Nicholas King, Scott~Mayer McKinney, Dean Carignan, and Eric Horvitz.
\newblock Capabilities of gpt-4 on medical challenge problems.
\newblock {\em arXiv preprint arXiv:2303.13375}, 2023.

\bibitem{touvron2023llama}
Hugo Touvron, Louis Martin, Kevin Stone, Peter Albert, Amjad Almahairi, Yasmine Babaei, Nikolay Bashlykov, Soumya Batra, Prajjwal Bhargava, Shruti Bhosale, et~al.
\newblock Llama 2: Open foundation and fine-tuned chat models.
\newblock {\em arXiv preprint arXiv:2307.09288}, 2023.

\bibitem{hu2021lora}
Edward~J Hu, Yelong Shen, Phillip Wallis, Zeyuan Allen-Zhu, Yuanzhi Li, Shean Wang, Lu~Wang, and Weizhu Chen.
\newblock Lora: Low-rank adaptation of large language models.
\newblock {\em arXiv preprint arXiv:2106.09685}, 2021.

\bibitem{li2021prefix}
Xiang~Lisa Li and Percy Liang.
\newblock Prefix-tuning: Optimizing continuous prompts for generation.
\newblock {\em arXiv preprint arXiv:2101.00190}, 2021.

\bibitem{lester2021power}
Brian Lester, Rami Al-Rfou, and Noah Constant.
\newblock The power of scale for parameter-efficient prompt tuning.
\newblock {\em arXiv preprint arXiv:2104.08691}, 2021.

\bibitem{n2c2portal}
Department of~Biomedical Informatics~at Harvard Medical~School.
\newblock n2c2 nlp research data sets, 2021.

\bibitem{pmlr-v139-zhao21c}
Zihao Zhao, Eric Wallace, Shi Feng, Dan Klein, and Sameer Singh.
\newblock Calibrate before use: Improving few-shot performance of language models.
\newblock In Marina Meila and Tong Zhang, editors, {\em Proceedings of the 38th International Conference on Machine Learning}, volume 139 of {\em Proceedings of Machine Learning Research}, pages 12697--12706. PMLR, 18--24 Jul 2021.

\bibitem{liu-etal-2022-makes}
Jiachang Liu, Dinghan Shen, Yizhe Zhang, Bill Dolan, Lawrence Carin, and Weizhu Chen.
\newblock What makes good in-context examples for {GPT}-3?
\newblock In Eneko Agirre, Marianna Apidianaki, and Ivan Vuli{\'c}, editors, {\em Proceedings of Deep Learning Inside Out (DeeLIO 2022): The 3rd Workshop on Knowledge Extraction and Integration for Deep Learning Architectures}, pages 100--114, Dublin, Ireland and Online, May 2022. Association for Computational Linguistics.

\bibitem{reimers-2019-sentence-bert}
Nils Reimers and Iryna Gurevych.
\newblock Sentence-bert: Sentence embeddings using siamese bert-networks.
\newblock In {\em Proceedings of the 2019 Conference on Empirical Methods in Natural Language Processing}. Association for Computational Linguistics, 11 2019.

\bibitem{peft}
Sourab Mangrulkar, Sylvain Gugger, Lysandre Debut, Younes Belkada, Sayak Paul, and Benjamin Bossan.
\newblock Peft: State-of-the-art parameter-efficient fine-tuning methods.
\newblock \url{https://github.com/huggingface/peft}, 2022.

\bibitem{alsentzer2019publicly}
Emily Alsentzer, John~R Murphy, Willie Boag, Wei-Hung Weng, Di~Jin, Tristan Naumann, and Matthew McDermott.
\newblock Publicly available clinical bert embeddings.
\newblock {\em arXiv preprint arXiv:1904.03323}, 2019.

\bibitem{yang2022large}
Xi~Yang, Aokun Chen, Nima PourNejatian, Hoo~Chang Shin, Kaleb~E Smith, Christopher Parisien, Colin Compas, Cheryl Martin, Anthony~B Costa, Mona~G Flores, et~al.
\newblock A large language model for electronic health records.
\newblock {\em NPJ Digital Medicine}, 5(1):194, 2022.

\bibitem{zhao2023dynamic}
Fei Zhao, Taotian Pang, Zhen Wu, Zheng Ma, Shujian Huang, and Xinyu Dai.
\newblock Dynamic demonstrations controller for in-context learning.
\newblock {\em arXiv preprint arXiv:2310.00385}, 2023.

\bibitem{reynolds2021prompt}
Laria Reynolds and Kyle McDonell.
\newblock Prompt programming for large language models: Beyond the few-shot paradigm.
\newblock In {\em Extended Abstracts of the 2021 CHI Conference on Human Factors in Computing Systems}, pages 1--7, 2021.

\bibitem{min2022rethinking}
Sewon Min, Xinxi Lyu, Ari Holtzman, Mikel Artetxe, Mike Lewis, Hannaneh Hajishirzi, and Luke Zettlemoyer.
\newblock Rethinking the role of demonstrations: What makes in-context learning work?
\newblock {\em arXiv preprint arXiv:2202.12837}, 2022.

\bibitem{lu2021fantastically}
Yao Lu, Max Bartolo, Alastair Moore, Sebastian Riedel, and Pontus Stenetorp.
\newblock Fantastically ordered prompts and where to find them: Overcoming few-shot prompt order sensitivity.
\newblock {\em arXiv preprint arXiv:2104.08786}, 2021.

\bibitem{zhou2022large}
Yongchao Zhou, Andrei~Ioan Muresanu, Ziwen Han, Keiran Paster, Silviu Pitis, Harris Chan, and Jimmy Ba.
\newblock Large language models are human-level prompt engineers.
\newblock {\em arXiv preprint arXiv:2211.01910}, 2022.

\end{thebibliography}
\end{document}